\documentclass[journal]{IEEEtran}

\ifCLASSINFOpdf
\else
\fi

\usepackage{amsmath}
\usepackage{amsfonts}
\usepackage{mathtools}

\def\Vec#1{\mbox{\boldmath $#1$}}

\usepackage{booktabs}
\usepackage{multirow}
\usepackage[T1]{fontenc}
\usepackage{comment}
\usepackage{graphicx}

\begin{document}
\title{Statistical Deformation Reconstruction Using Multi-organ Shape Features for Pancreatic Cancer Localization}

\author{Megumi~Nakao,~\IEEEmembership{Member,~IEEE,}
        Mitsuhiro~Nakamura,
        Takashi~Mizowaki
        and~Tetsuya~Matsuda,~\IEEEmembership{Member,~IEEE}
\thanks{M. Nakao, and T. Matsuda are with Graduate School of Informatics, Kyoto University, Yoshida-Honmachi, Sakyo, Kyoto, 606-8501, JAPAN, e-mail: megumi@i.kyoto-u.ac.jp.}
\thanks{M. Nakamura is with Graduate School of Medicine, Human Health Sciences, Kyoto University.}
\thanks{T. Mizowaki is with Dept. of Radiation Oncology and Image-Applied Therapy, Kyoto University Hospital.}
\thanks{M. Nakao and M. Nakamura are contributed equally to this work.}}

\markboth{Journal of \LaTeX\ Class Files,~Vol.~, No.~, Nov~2019}%
{Nakao \MakeLowercase{\textit{et al.}}: Bare Demo of IEEEtran.cls for IEEE Journals}

\maketitle

\begin{abstract}
    Respiratory motion and the associated deformations of abdominal organs and tumors are essential information in clinical applications. However, inter- and intra-patient multi-organ deformations are complex and have not been statistically formulated, whereas single organ deformations have been widely studied. In this paper, we introduce a multi-organ deformation library and its application to deformation reconstruction based on the shape features of multiple abdominal organs. Statistical multi-organ motion/deformation models of the stomach, liver, left and right kidneys, and duodenum were generated by shape matching their region labels defined on four-dimensional computed tomography images. A total of 250 volumes were measured from 25 pancreatic cancer patients. This paper also proposes a per-region-based deformation learning using the reproducing kernel to predict the displacement of pancreatic cancer for adaptive radiotherapy. The experimental results show that the proposed concept estimates deformations better than general per-patient-based learning models and achieves a clinically acceptable estimation error with a mean distance of 1.2 $\pm$ 0.7 mm and a Hausdorff distance of 4.2 $\pm$ 2.3 mm throughout the respiratory motion. 
\end{abstract}


\IEEEpeerreviewmaketitle

\section{Introduction}

\begin{figure}[t]
	\begin{center}
    \includegraphics[width=85mm]{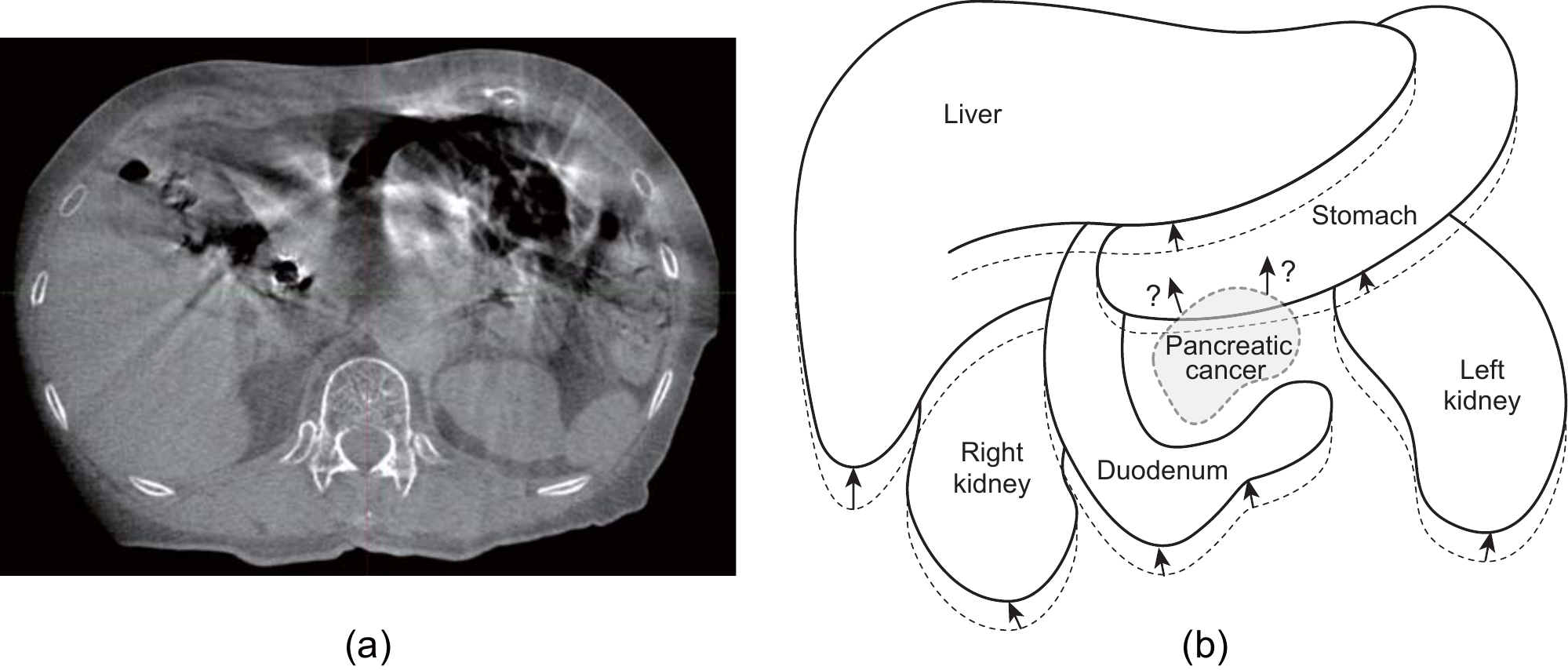}
    \hspace{-1cm}
  \end{center}
	\caption{Clinical needs for indirect deformation reconstruction in ART: (a) CBCT images with missing pixels and (b) a conceptual image for pancreatic cancer localization from multi-organ shape features.}
	\label{fig:1}
\end{figure}

\IEEEPARstart{S}{tatistical} formulation of respiratory motion including the deformation of multiple organs and tumors is of increasing interest in external-beam radiotherapy. Specifically, in image-guided radiotherapy (IGRT), it is important to reduce exposure in normal tissues while accurately targeting lesions with respiratory motion. However, large anatomical variations in the shape and motion of abdominal organs can occur when treatment lasts several weeks, while time-series three-dimensional (3D) computed tomography (CT) images can only be obtained for initial radiation planning \cite{Rigaud19}\cite{Magallon18}. Recent technical advances enable treatment plans to be modified based on a daily X-ray cone-beam CT (CBCT). This process is called adaptive radiotherapy (ART) \cite{Posiewnik19}\cite{Hvid18}. Abdominal organs such as the stomach, and duodenum or the pancreas neighbor each other, but the pancreas cannot clearly be detected, even on CBCT images. Because missing pixel values or artifacts often appear in CBCT images (see Fig. \ref{fig:1}), ART for abdominal regions is technically more difficult and remains a challenging area of research \cite{Magallon18}\cite{Posiewnik19}. To approach these issues, this paper focuses on the potential of model-based tumor localization in pancreatic cancer treatment using statistical modeling of patient-specific multi-organ motion and deformation.

Statistical shape modeling (SSM) has been widely investigated for the modeling of organ shapes based on prior knowledge \cite{Heimann09}, and interest has grown in its application in machine learning with intra- or inter-patient datasets with point-to-point correspondence \cite{Rigaud19}\cite{Magallon18}\cite{Nakamura19}\cite{Tilly17}. For instance, interfractional shape variations in the prostate and rectum  have been statistically modeled for radiation therapy planning \cite{Nakamura19}\cite{Hekal18}\cite{Bondar14}. A statistical deformation model (SDM) \cite{Ehrhardt11}\cite{Jud17} based on a four-dimensional (4D) CT images has also been reported. Unlike physics-based modeling \cite{Fuerst15}\cite{Nakao10}\cite{Nakao07}, statistical modeling is a data-driven approach that does not explicitly describe the elasticity and physical conditions of organs. Image-based \cite{Sotiras13}\cite{Ruhaak17}, point-based \cite{Hekal18}\cite{Shibayama17}, or mesh-based \cite{Rigaud19}\cite{Magallon18}\cite{Nakamura19}\cite{Nakao19} deformable registration techniques have been used to obtain point-to-point correspondence between two datasets. Specifically, respiratory motion \cite{Jud17}\cite{Fuerst15}\cite{Ruhaak17}\cite{Wilm16} has mainly been investigated in the field of image-based lung modeling. 

Our aim in this paper is to statistically model inter- and intra-patient deformation along with the motion of multiple abdominal organs. The relationships among organs, especially those between pancreatic cancer and its surrounding organs, are also interesting and worth investigating \cite{Yu16}\cite{Fontana16}. Image-based registration assumes that the 3D image or the regions of interest of organs are a continuous space. It also assumes that the displacement is spatially smooth for regularization \cite{Sotiras13}\cite{Zhang18}\cite{Oh17}\cite{Klein10}. However, the mechanism for multi-organ deformation is complex and not fully understood. Respiratory motion widely affects abdominal regions, in which several organs are adjacent to or connected with each other. Therefore, to analyze individual motion and deformation accurately, a model capable of capturing sliding motion or rotation near the organ boundary is needed \cite{Jud17}. Hence, interest in deformable mesh registration (DMR) has recently resurged. The inter- and intra-patient variations in cervix--uterus anatomy were recently modeled for radiation therapy planning \cite{Rigaud19}\cite{Tilly17}. The correlation of liver and pancreas tumor motion \cite{Robert17} and daily changes in the stomach, duodenum, and pancreas \cite{Magallon18} have been analyzed through DMR. To the best of our knowledge, there has been no report describing SDM-based deformation reconstruction or clarifying the relationship between the inter- and intra-patient deformations of multiple organs and tumors, even though the clinical needs for such modeling in radiotherapy and motion recognition are high \cite{Jadon14}\cite{Iwai17}\cite{Teske15}\cite{Whitfield12}.

In this paper, we introduce a localized deformation reconstruction framework based on the shape features of multiple organs. To model the relationships of the spatial deformation fields between abdominal organs and the tumor, a statistical multi-organ deformation model (SMDM) is constructed by shape matching organ meshes generated from 4D-CT datasets (250 volumes). We formulate the SMDM as a reproducing kernel, with the motion and deformation of the pancreatic cancer region as the estimation target. Five abdominal organs (the liver, stomach, duodenum, and right and left kidneys) located around the pancreas are used as multidimensional shape features. In such a procedure, a low number of cancer patients in the dataset can decrease the estimation performance, which is a common problem in clinical machine learning. This paper proposes the concept of localized deformation reconstruction to model nonlinear motion for a local small region of the estimation target, rather than on a per-patient basis. Here, to design the reproducing kernel framework that maps motion with deformation of pancreatic cancer and its surrounding organs, we address the following fundamental issues: 

\begin{itemize}
    \item We investigate the level of complexity of the motion dynamics of multiple abdominal organs, their deformation, and relationships. 
    \item We determine the design of DMR to build SMDM in abdominal region. 
    \item We investigate which organ sets are good estimators to predict the motion/deformation of gross tumor volume (GTV) and determine an appropriate number of dimensions of the feature space.
    \item We evaluate the final estimation performance for the inter- and intra-patient validation of GTV and address whether the proposed concept is clinically acceptable.
\end{itemize}

We note that the purpose of shape/deformation reconstruction is different from that of image segmentation, for which a variety of deep learning methods that directly use image features in the regions of interest have been proposed  \cite{Xu19}.  In the proposed kernel-based formulation of SMDM, as shown in Fig. \ref{fig:1}(b), even if pancreatic cancer is totally "invisible" because of missing pixels or artifacts in CBCT images, its motion and deformation can be indirectly reconstructed from multi-organ features. This estimation helps optimize radiation treatment in that the radiation dose can be locally transported to the moving tumor while keeping safe a margin around organs at risk. 


\begin{figure*}[t]
	\begin{center}
		\includegraphics[width=180mm]{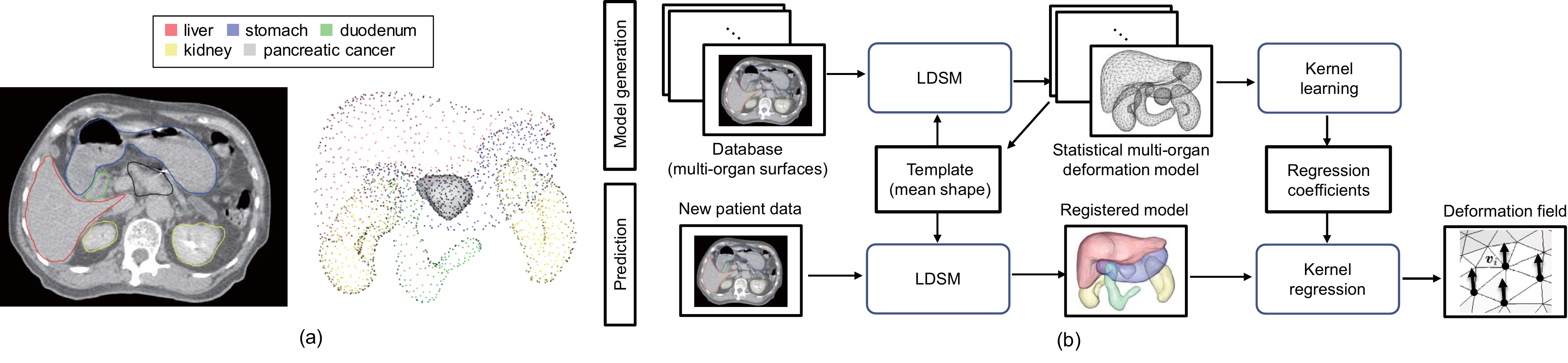}
	\end{center}
	\caption{Deformation reconstruction framework using the shape features of multiple abdominal organs. (a) CT slice image with the 3D contours and vertex distribution of the meshes. (b) The template meshes are registered to individual organ regions using Laplacian-based diffeomorphic shape matching (LDSM). The methods are mainly used to predict the motion and deformation of pancreatic cancer using reproducing kernel regression. }
	\label{fig:3}
\end{figure*}

\section{Methods}

\subsection{Statistical Multi-organ Deformation Model}
In this study, 4D-CT images ${I}_t^{(k)}$ ($k=1, 2, ..., 25$: patient ID, $t=1, 2, ..., 10$: time phase of 4D-CT images) of 25 pancreatic cancer patients who underwent intensity-modulated radiotherapy (IMRT) in Kyoto University Hospital were used for statistical modeling and deformation learning. This study was performed in accordance with the Declaration of Helsinki and was approved by our institutional review board (approval number: R1446). 4D-CT images for a patient consisted of 3D-CT images (image size: $512 \times 512$ pixels and 88--152 slices, voxel resolution: $1.0 mm \times 1.0 mm \times 2.5mm $) of 10 time phases for one respiratory cycle. All images were measured under the condition of respiratory synchronization, where $t=1$ corresponds to the end-inhalation phase and $t=6$ corresponds to the end-exhalation phase. The region labels of the entire body, stomach, liver, duodenum, left and right kidneys, and the GTV of the pancreatic cancer were predefined by board-certified radiation oncologists in the radiotherapy planning, as shown in Fig. \ref{fig:3}(a). We note that accurate automatic extraction is difficult for the stomach, duodenum, and pancreas because of their unclear boundaries, contrasts, and air contents. Their shapes can be substantially deformed during the radiotherapy period. Most of anatomical segmentation techniques assume that there are no image defects nor a wide range of artifacts appearing in the CBCT images. In this study, the 3D contours manually defined as the organs at risk (OARs) for dose calculation in the radiation planning were used as the multi-organ shape database, and their tetrahedral meshes ${S}_t^{(k)}$ were generated. Figure \ref{fig:3}(a) show an example of tetrahedral meshes and their volumetric point distribution for five organs and the GTV of the pancreatic cancer. 

\begin{figure}[t]
  \centering
  \includegraphics[width=84mm]{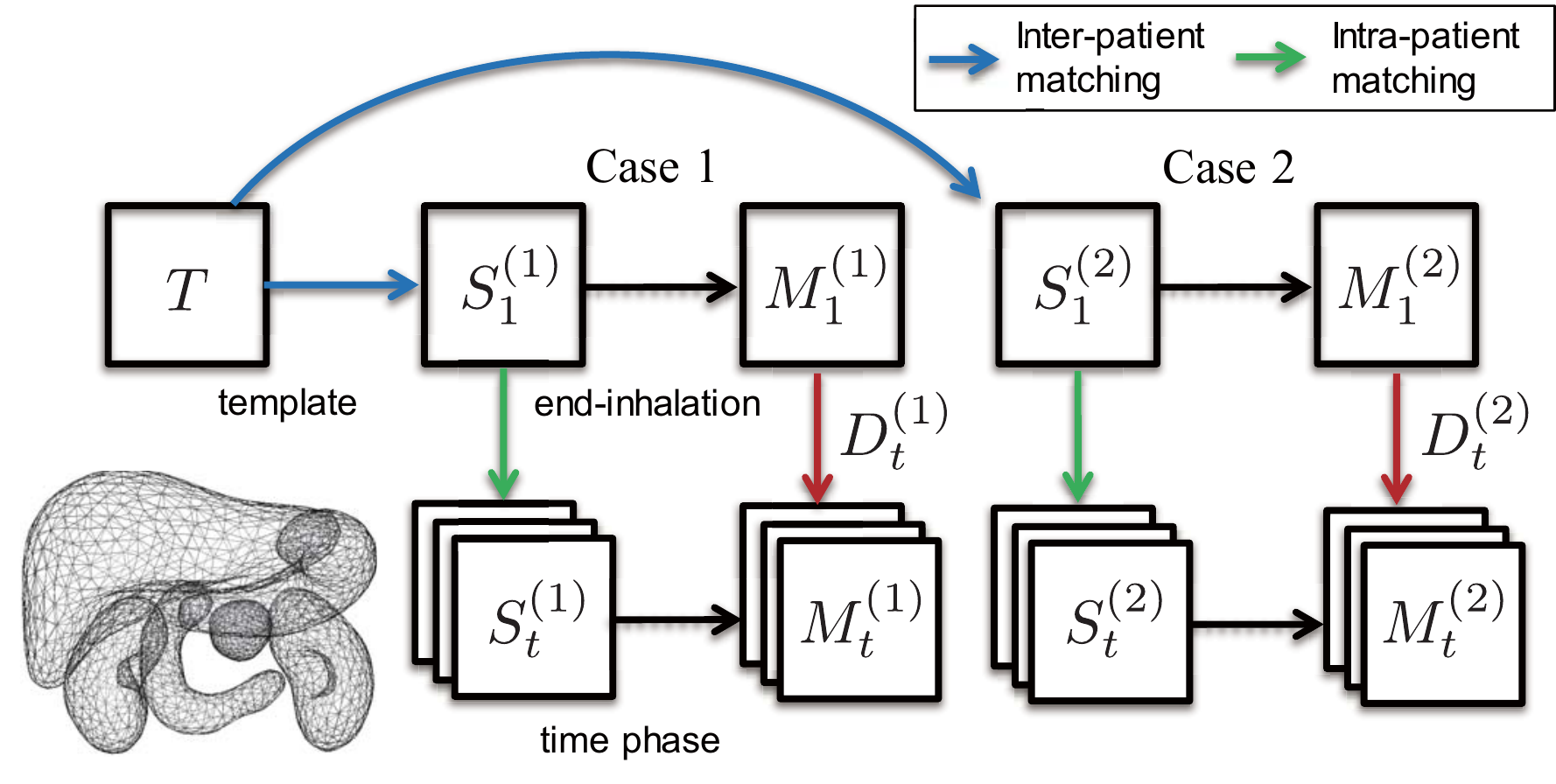}
  \caption{Flow of the inter- and intra-patient shape matching. The spatial deformation $D_{t}$ between the end-inhalation state and time phase $t$ is obtained from the corresponding vertices of the two registered meshes. }
  \label{fig:4}
\end{figure}

Figure \ref{fig:3}(b) shows the flow of the developed framework. The meshes $S_{t}^{(k)}$ differ in the number of vertices and the structure of the mesh, because they were independently generated from different CT images. As shown in Fig. \ref{fig:4}, the corresponding models $M_{t}^{(k)}$ (with the same vertex and the same mesh structure) that precisely approximate the surfaces of $S_{t}^{(k)}$, respectively, were computed by DMR using a template $T$ and the target mesh. Because the registered models achieve point-to-point correspondence, spatial deformation $D_{t}^{(k)} = M_{t}^{(k)} - M_{1}^{(k)}$ can be represented by calculating the displacement vector of the corresponding vertex. 

\begin{figure}[t]
  \centering
  \includegraphics[width=88mm]{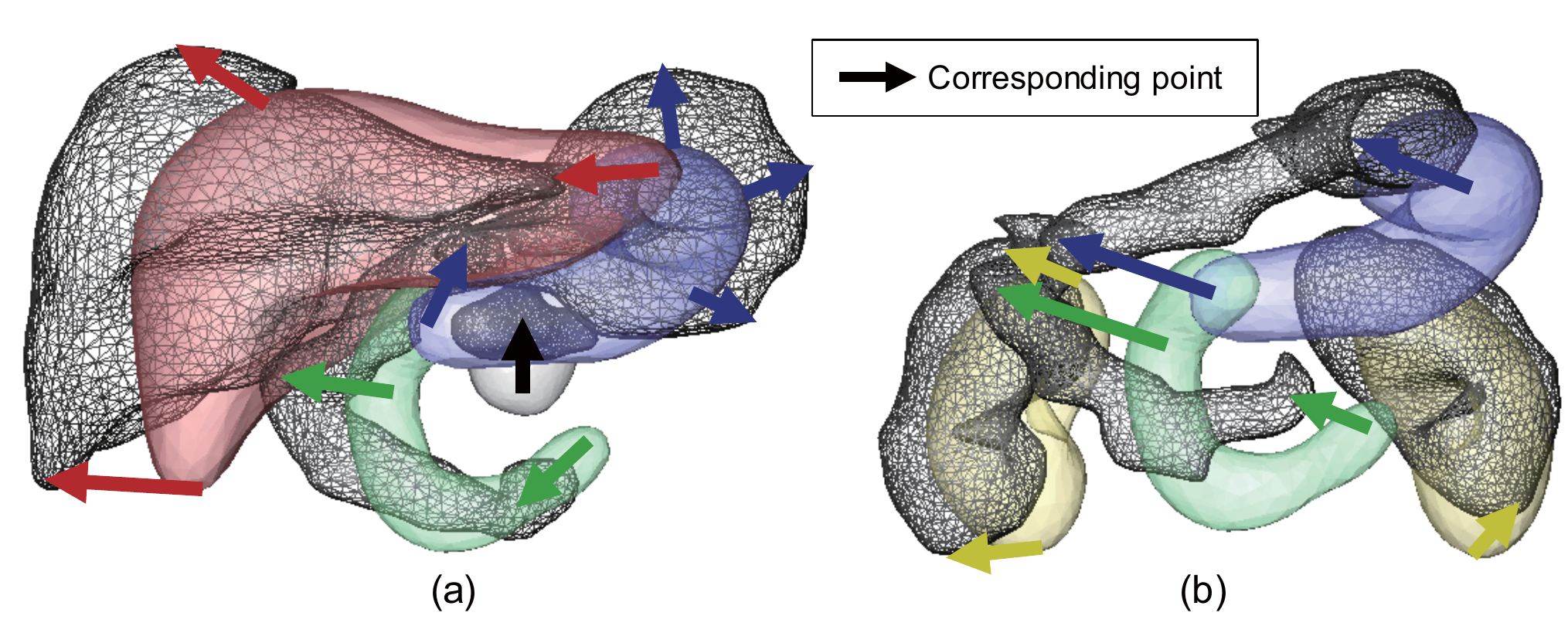}
  \caption{Examples of complex inter-patient shape changes and the interactions among multiple abdominal organs. (a) A part of the liver and stomach deform in different directions near their boundaries (red and blue arrows). (b) The duodenum and right kidney do not contact each other in the template model, but are in contact in the patient models (green and yellow arrows).}
  \label{fig:5}
\end{figure}

To stably compute a 3D deformation field of multiple organs, accurate shape matching is required. The shape variation and deformation of abdominal organs such as the stomach and duodenum include considerable volume changes including rotations and sliding boundaries. Figure \ref{fig:5} shows examples of complex inter-patient shape changes and the interaction between multiple abdominal organs. A part of the liver and stomach deform in the different directions near their boundaries. The duodenum and right kidney do not contact each other in the template model, but are in contact in the patient models. Image-based registration generally deals with multiple organs as one continuous image space, and an individual organ's deformation or interactions among specific organs cannot be discriminated.  In addition, the registration error tends to increase in areas with large curvature such as the tips of the organs and boundaries of neighboring organs \cite{Jud17}\cite{Nakao19}. Specifically, our focus in this paper is to model multi-organ motion/deformation and its interaction, and to examine the efficacy of the proposed feature-preserving registration methods for abdominal organs with large shape variations such as the stomach and duodenum. To capture rotational components or possible sliding motions caused by the interaction of multiple organs, the registration process is applied to each organ's mesh independently. In addition to intra-patient registration, our approach enables the construction of a SDM, making inter-patient deformation analysis possible. The details of the applied DMR method, LDSM, are described in Section II-C.

For the template mesh generation, first, one case was randomly selected to be the initial mesh for $T$, and its surface was resampled to 400 vertices and 796 triangles. Next, the corresponding mesh models ${M}_{I}^{(k)}$ were obtained by registering $T$ to the individual surfaces $S_{I}^{(k)}$ in the inflated state. Because the mesh models ${M}_{t}^{(k)}$ have point-to-point correspondence, the average shape $\overline{M}$ can be obtained by calculating the average of each coordinate. We used $\overline{M}$ as the final template. This process was performed on the stomach, liver, left and right kidneys, and duodenum. By keeping the template close to the data in advance, our aim is to reduce the influence of the adopted data selection while preventing an increase in matching error.

\subsection{Localized deformation learning model}

In the prediction stage, the time-varying position of the pancreatic cancer is estimated using the reproducing kernels of the multi-organ features. Here, as the second technical challenge, we introduce the localized multi-organ features for modeling inter- and intra-patient variation of deformation from a dataset containing a limited number of patients. To model spatial deformation, per-patient-based learning, in which one patient's data are used as one set of training data, has been generally employed as a straightforward approach. The displacement of the region is computed by synthesizing that of the corresponding regions of the training dataset with similar shape features. In this case, the number of training data equals the number of patients in dataset $m$. 

Instead of reconstruction on a per-patient basis, the aim of this paper is to formulate a method for local deformation reconstruction that models nonlinear motion for a small region of the estimation target. In the proposed learning model, a small local region of the target organ is used as one set of training data. This approach is based on the hypothesis that local regions with similar shape features show similar displacements, which is commonly assumed in each continuous space of organs. In this approach, the displacement of the region is calculated by combining the training datasets of all the regions including other parts of the target organs. The number of training data is $m \times n$ when the number of patients in the data is $m$ and $n$ small regions of the target organ are considered. The details of the deformation reconstruction framework is described in Section II-D.

\subsection{Feature preserving deformable mesh registration}
In this study, to achieve both globally stable and locally strict DMR, our aim is to address the trade-off between feature-preserving shape matching and spatially smooth deformation. To approach this in DMR, the concept of progressive, feature-preserving shape update \cite{Nakao19}\cite{Kim15} is introduced into the large deformation diffeomorphic metric mapping (LDDMM) scheme \cite{Zhang18}\cite{Faisal05}. The objective function in the proposed LDSM is described as follows.

\begin{eqnarray}
    E(\Vec{u}) = d(Y, \phi(X)) + \int_{0}^{1} \|L(\Vec{u}(s))\|^2 ds,  
\end{eqnarray}
where $X$ is the source (template) mesh, $Y$ is the target surface, and $d$ is the distance function between the two surfaces. In addition, $\phi(X)$ is a continuous and differentiable transformation that maps $X$ to the deformed mesh, $L(\cdot)$ is the Laplace--Beltrami operator and $L(\Vec{u})$ is the discrete Laplacian of the displacement field.   

The first term evaluates the difference between the deformed template and the target surface. The second is a regularization term to make the deformation field \Vec{u} smooth. In the context of image-based LDDMM, the difference in voxel intensities between the deformed and the target image is evaluated in the first term. Minimizing the nearest point-to-point distance was originally employed as a basic strategy in DMR, however, it does not consider 3D geometry, and maintaining mesh topology is difficult with this approach. We focus on the importance of feature preservation to avoid incorrectly matching local structures \cite{Nakao19} in mapping distant structures, 3D geometric information of the template mesh is used to preserve the local features of organ shapes. In the mapping function $\phi$, a Laplacian-based shape matching (LSM) scheme \cite{Kim15} is introduced for progressive shape updates while preserving features as much as possible. The mean value of the nearest bidirectional point-to-surface distance (called the mean distance in this paper) is used for $d$. A discrete Laplacian was first formulated for geometry modeling \cite{Nealen06}, and has recently been applied to the non-rigid shape registration of anatomical structures \cite{Nakao19}\cite{Kim15}. In \cite{Kim15}, LSM registers curved surfaces with shape variations better than LDDMM. 

Let $X$ denote a tetrahedral mesh with $n$ vertices $\Vec{v}_{i} \in V (i=1, 2, ..., n)$ and edges, the deformation map $\phi(X)$, that is, the deformation field $\Vec{u}$ in the LSM, is obtained by iteratively updating $\Vec{v}_{i}$ while minimizing the following objective function:

\begin{align}
    E(\Vec{u}) &= E_{shape} + \delta E_{pos} \\ \nonumber
            &= \sum_{i=1}^{n} \| L( \Vec{v}'_{i})\ - L(\Vec{v}_{i}) \|^{2} + \delta \sum_{i =1}^{n} \|\Vec{p}_{i} - \Vec{v}_{i} \|^{2},
\end{align}
where ${\Vec v'_i}$ is the vertex position to be solved, ${\Vec p_i}$ is a positional constraint set to ${\Vec v_i}$, and $\delta$ is a weight parameter configured according to the problem. For the definition of the positional constraints between the template and target shape, refer to \cite{Nakao19}. $L({\Vec v_i})$ is the discrete Laplacian at vertex $ {\Vec v_i} $, defined by

\begin{eqnarray}
	L(\Vec v_{i}) = \sum_{j \in N(\Vec v_{i})} \omega_{ij} (\Vec v_{i}- \Vec v_{j})
\end{eqnarray}
Here, $\omega_{ij}$ are the edge weights and $N({\Vec v_i})$ is the number of adjacent vertices of one ring connected by vertex ${\Vec v_i}$ and the edge. The discrete Laplacian is used as a shape descriptor and approximates the mean curvature normal of the triangular mesh. Although there are several variations of the weights, the general one is cotangent discretization based on the per-edge Voronoi areas. The first term in Eq. (2) is a penalty for shape changes to the mesh, and the second term increases if the constrained vertex is distant from the nearest surface of the target mesh. By computing ${\Vec v'_i}$, which minimizes the objective function, the template model is updated while preserving the shape as much as possible. 

%

In the proposed LDSM, in addition to LSM, the regularization term in Eq. (1) is used to achieve a spatially smooth and diffeomorphic deformation. This can be simply implemented by adding the term to Eq. (2) as follows:

\begin{align}
    E(\Vec{u}) &= E_{shape} + E_{deform} + \delta E_{pos} 
\end{align}
where $E_{deform}$ is the magnitude of the discrete Laplacian of the deformation field $\Vec{u}$ defined by

\begin{align}
    E_{deform} &= \sum_{i=1}^{n} \|L(\Vec{u}_{i})\|^{2}
\end{align}

Because Eq. (5) is a quadratic form at vertex positions ${\Vec v_i}$, the minimization problem can be calculated efficiently. The step-by-step update avoids local mismatches at the early stage if there is a considerable distance between the two surfaces. Once the template is updated, local displacement $\Vec{u}_{i} = \Vec{v}'_{i} - \Vec{v}_{i}$ is obtained. We note that the original LDDMM does not preserve the shape of the template, and some studies report results with unstable or irregular matching especially for large deformations with rotation \cite{Nakao19}\cite{Kim15}. In the proposed framework, in addition to the regularization term, Laplacian-based shape preservation is introduced into the iterative deformation process. 

\subsection{Reproducing kernels for deformation reconstruction}
\begin{figure}[t]
    \centering
    \includegraphics[width=84mm]{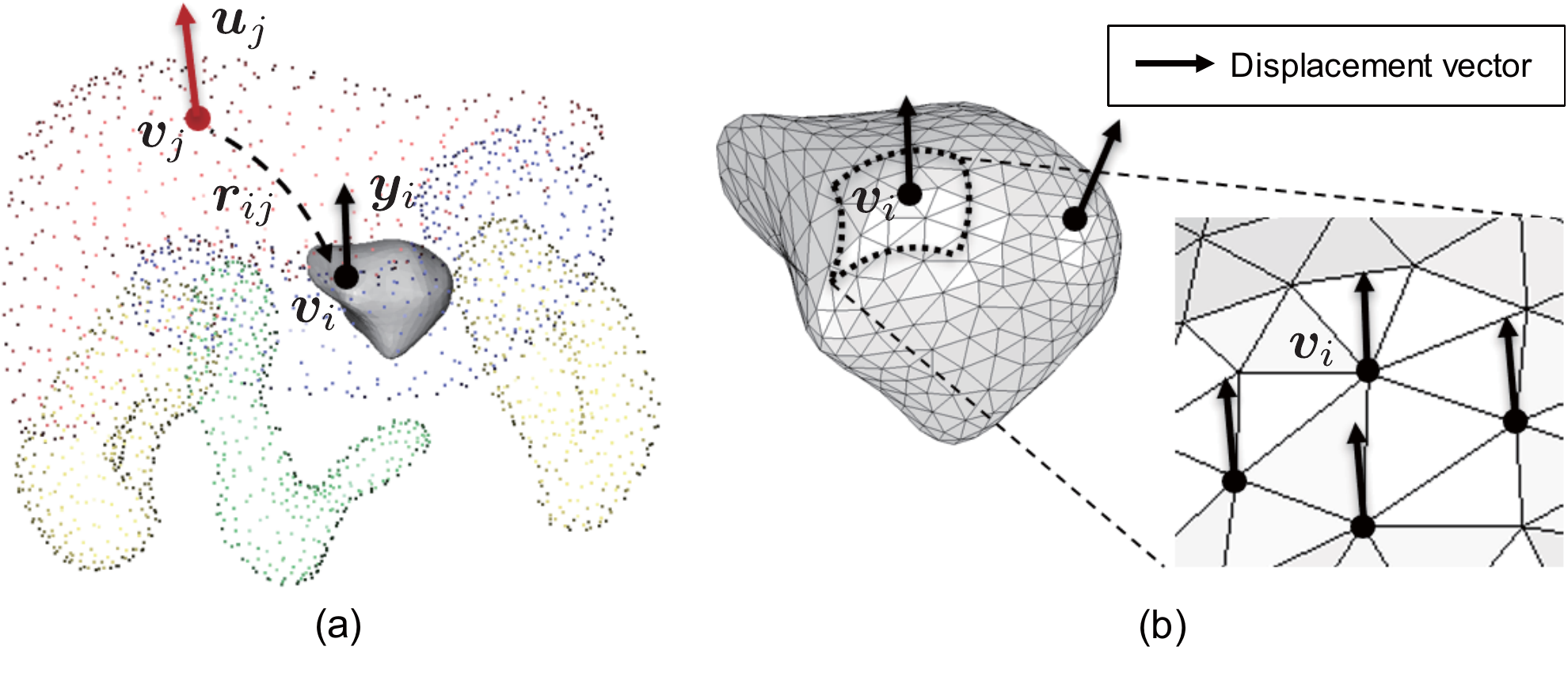}
    \caption{Per-region-based localized deformation learning: (a) a displacement mapping model, and (b) deformation learning by considering the continuity of spatial deformation. }
    \label{fig:6}
\end{figure}

This section explains the per-region-based localized deformation learning proposed in this paper, by comparing it with the per-patient-based deformation learning generally used in population-based modeling \cite{Rigaud19}\cite{Nakamura19}. Figure \ref{fig:6} briefly illustrates the concept. Because the aim of this paper is to reconstruct target deformation from multi-organ features, the displacement vector $\Vec{y}_{i}$ is mapped from multiple points sampled from the surrounding organs, as shown in Fig. \ref{fig:6}(a). The local displacement ${\Vec y}_{i}$ of GTV can be calculated from the feature vectors of the sampled points and is composed using the relative position ${\Vec r}_{ij}$ and displacement ${\Vec u}_{j}$. 

Here, the problem is to learn the mapping from the registered multi-organ models generated through DMR. Per-patient-based learning is a straightforward approach, where the mesh model $X$ obtained from one patient's data is used as one set of training data. It is based on the idea that the displacement of local regions of organs is similar to that of the corresponding regions of the training dataset with similar shape features. Specifically, the displacement $\Vec{u}_i$ at vertex $i$ is learned from the displacement $\Vec{u}_i$ of the corresponding vertex $\Vec{v}_i $ in other cases.  However, in per-region-based localized deformation learning, a small region $\Omega$ of the mesh model $X$ is used as one set of training data. This approach is based on the idea that local regions with similar shape features show similar displacement, which is commonly assumed if a curved manifold is used to represent an organ surface. For instance, in Fig. \ref{fig:6}(b), the displacement vectors adjacent to $\Vec{v}_{i}$ should be similar to those of $\Vec{v}_{i}$. In other words, the displacement $\Vec{u}_i $ can be learned from the displacement $\Vec{u}_j$ of all vertices $\Vec{v}_j$ in other cases.

\begin{figure}[t]
    \centering
    \includegraphics[width=76mm]{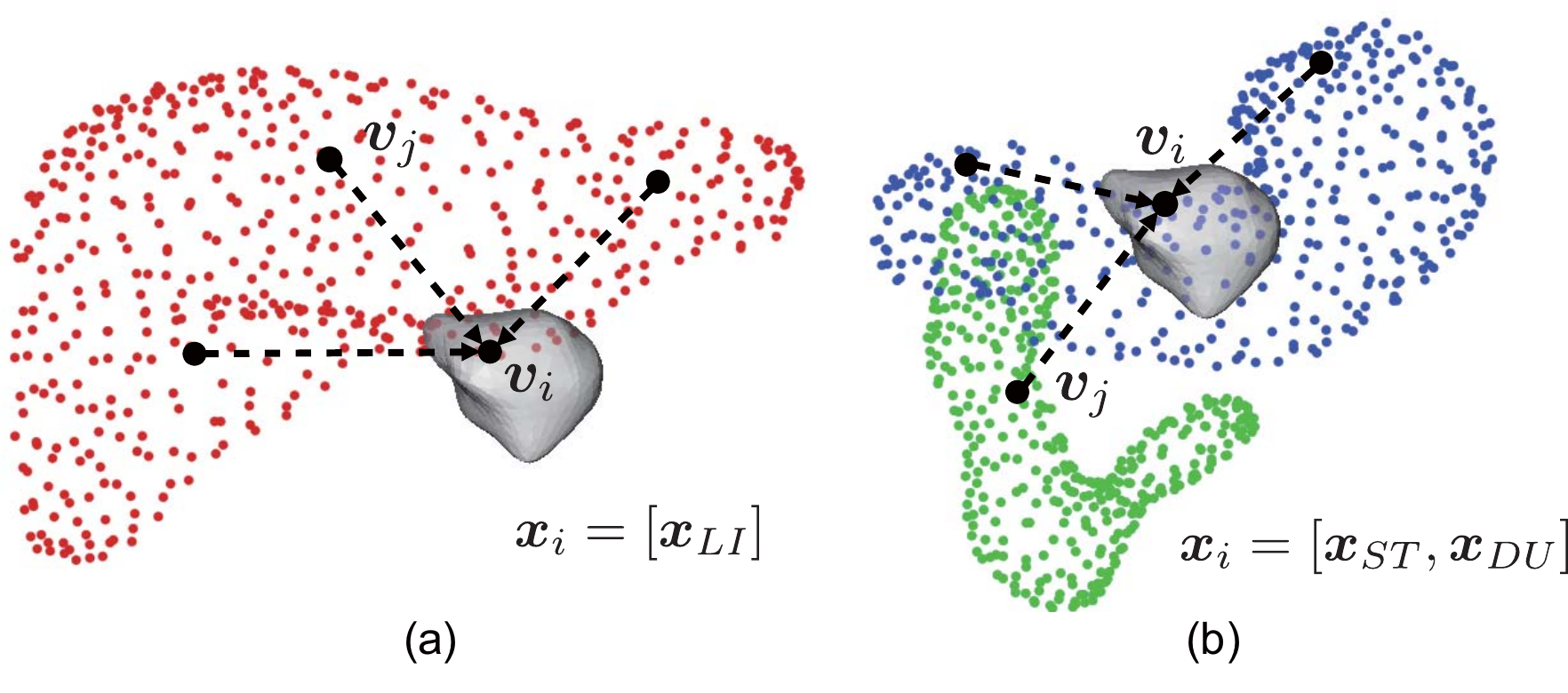}
    \caption{Examples of displacement mapping models using multi-organ shape features. (a) Shape features sampled from the liver and (b) from the stomach and duodenum. }
    \label{fig:7}
\end{figure}

We formulate the per-region-based deformation learning model using reproducing kernels. Based on the spatial mapping in Fig. \ref{fig:6}(a), the local displacement $\Vec{y}_{i}$ of GTV is computed from multi-organ shape features $\Vec{x}_{i}$ using Eq. (\ref{eq:reproducing_kernel}).

\begin{eqnarray}
\Vec{y}_{i} = \sum_{j=1}^{N} \Vec{\alpha}_j k(\Vec{x}_{i},\Vec{x}_j), \hspace{.3cm} \Vec{x}_{j}\in \it{X}, \hspace{.3cm}\Vec{\alpha}_{j}\in\mathbb{R}^{N}
\label{eq:reproducing_kernel}
\end{eqnarray}
where $k :\it{X} \times \it{X}\rightarrow\mathbb{R}$ is the reproducing kernel, $N$ is the number of training datasets, and $\Vec{\alpha}_{j}$ is the weight vector. A Gaussian function is used for kernel function $k$, which is $k(\Vec{x}_i,\Vec{x}_j) = exp(- \beta ||\Vec{x}_i - \Vec{x}_j||^2 / N)$. For a given $\Vec y = [\Vec{y}_1, \dots, \Vec{y}_N]^T$, $\Vec \alpha = [\Vec{\alpha}_1, \dots,\Vec{\alpha}_N]^T$ are calculated by minimizing the cost function E(\Vec{\alpha}), which is expressed as

\begin{eqnarray}
E(\Vec{\alpha})	= \| \Vec{y} - K \Vec{\alpha} \|^{2} + \lambda \Vec{\alpha}^{T} K \Vec{\alpha}
\label{alphaの定義}
\end{eqnarray}
where $K \in \mathbb{R}^{N \times N} $ is the kernel matrix whose elements are defined by $K_{ij} = k(\Vec{x}_{i},\Vec{x}_j)$, and $\lambda$ is the regularization parameter, which penalizes deviations of $\Vec{\alpha}$. The optimized weights are given by $\Vec \alpha = (K + \lambda I)^{-1} \Vec y$ ($I$: indentity matrix).

The feature vector $\Vec{x}_{i}$ of the local region for per-region-based learning is constructed using the relative position $\Vec{r}_{ij}$ of the target vertex and the displacement vector $\Vec{u}_j$ of the surrounding organs as follows.
  
\begin{eqnarray}
\Vec{x}_{i} = [\Vec{r}_{ij} , \Vec{u}_j], \hspace{.3cm} \Vec{r}_{ij} = \Vec{v}_i - \Vec{v}_j
\label{eq:feature_vector}
\end{eqnarray}

In this study, the sampled vertices of the SMDM obtained from 24 patients were used in leave-one-out cross-validation to construct the kernel matrix, and all vertices of the five organs (liver, left and right kidneys, stomach, and duodenum) were considered as candidates for shape features $\Vec{x}$. 

\begin{eqnarray}
    \Vec{x}_{ALL} = [\Vec{x}_{ST}, \Vec{x}_{DU}, \Vec{x}_{LI}, \Vec{x}_{LK}, \Vec{x}_{RK}]
    \label{eq:all_feature_sets}
\end{eqnarray}
where ST, DU, LI, LK, and RK are the stomach, duodenum, liver, left kidney, and right kidney, respectively. For instance, when 50 vertices are sampled from each organ model, the dimension of the feature vector $\Vec{x}_{i}$ is 1,500 according to $\Vec{r}_{ij}, \Vec{u}_j \in \mathbb{R}^{750}$. Because the shape features are evaluated in high-dimensional feature space, in per-patient-based deformation reconstruction, the estimation error may increase if there are no corresponding vertices with similar characteristics among the 24 cases.

Alternatively, in the proposed per-region-based deformation learning, the displacement is locally learned per vertex; in other words, it can be reconstructed from the deformation of different regions. Therefore, when 200 vertices are used to represent the target pancreatic cancer (for example), the training dataset is substantially increased to $24 \times 200 = 4,800$, whereas the number of dimensions of the feature space is 1,500, which is the same as that of per-patient-based approach. This means that the number of neighborhood data points increase in the generated feature space, and the displacement of the target vertex can be reconstructed using more datasets. Therefore, using a per-region-based kernel formulation improves the estimation performance and more stable results can be expected. 

As described in the introduction, we are here interested in the following questions: which organ sets are good estimators for deformation reconstruction and what number of dimensions of the feature space is appropriate. Figure \ref{fig:7} shows two examples in which the three vertices are mapped to the target vertex, which is a problem of obtaining the target displacement based on features sampled from different organ sets, $[\Vec{x}_{LI}]$ from the liver or $[\Vec{x}_{ST}, \Vec{x}_{DU}]$ from the stomach and duodenum. The liver does not have clear anatomical connectivity to the pancreas, and the shape feature is relatively stable with a small deformation. The stomach and duodenum are connected to the pancreas, but their deformation is large \cite{Magallon18} and unstable. Because the shape features of all organs are not always obtained from CBCT images, to explore the prediction performance when using specific organ sets is worth investigating for clinical application. In the experiments, we examine the efficacy of the proposed concept and investigate the possible organ sets and the proper number of dimensions of the feature space for localizing pancreatic cancer.

\section{Experiments}
In the experiments, statistical multi-organ deformation models were first generated by inter- and intra-patient shape matching. The registration performance for each organ was confirmed while comparing it with the results from three existing registration methods. Then, the efficacy of the multi-organ shape features in predicting the motion and deformation of pancreatic cancer was analyzed. The performance of the proposed per-region-based deformation learning was evaluated by comparing it with conventional per-patient-based learning. The value of weight parameter $\beta$ for kernel function was determined from the results of numerical experiments. 

\subsection{Shape matching performance}

In this study, the mean distance \cite{Rigaud19}\cite{Kim15}, the Hausdorff distance \cite{Huttenlocher93}, and Laplacian of the displacement \cite{Nakao19} were used as the shape similarity criteria. The Hausdorff distance measures the longest distance between two surfaces, whereas the mean distance is the mean value of the nearest bidirectional point-to-surface distance. Unlike segmentation or recognition problems, statistical modeling requires point-to-point local correspondence between two shapes. For example, because the Dice coefficient only measures volume overlap, it is not suitable for evaluating per-vertex correspondence, nor is it suitable for measuring the quality of local matching. The Laplacian of the displacement is the magnitude of the second derivatives of the displacement field, and it evaluates the smoothness of deformation. If local shape matching is achieved while preserving vertex density, this value decreases. 

The proposed shape matching method (LDSM)  was compared with three existing shape matching approaches, that is, 

\begin{itemize}
    \item Piecewise Affine Transformation (PWA) \cite{Pitiot06}\cite{Zhou10}
    \item Large deformation diffeomorphic metric matching \\ (LDDMM) \cite{Faisal05}
    \item Laplacian-based shape matching (LSM) \cite{Kim15}\cite{Saito15}
\end{itemize}

For all algorithms, the affine transformation was processed in advance to match the posture and volume of the overall shape globally.

\begin{figure}[t]
    \centering
    \includegraphics[width=80mm]{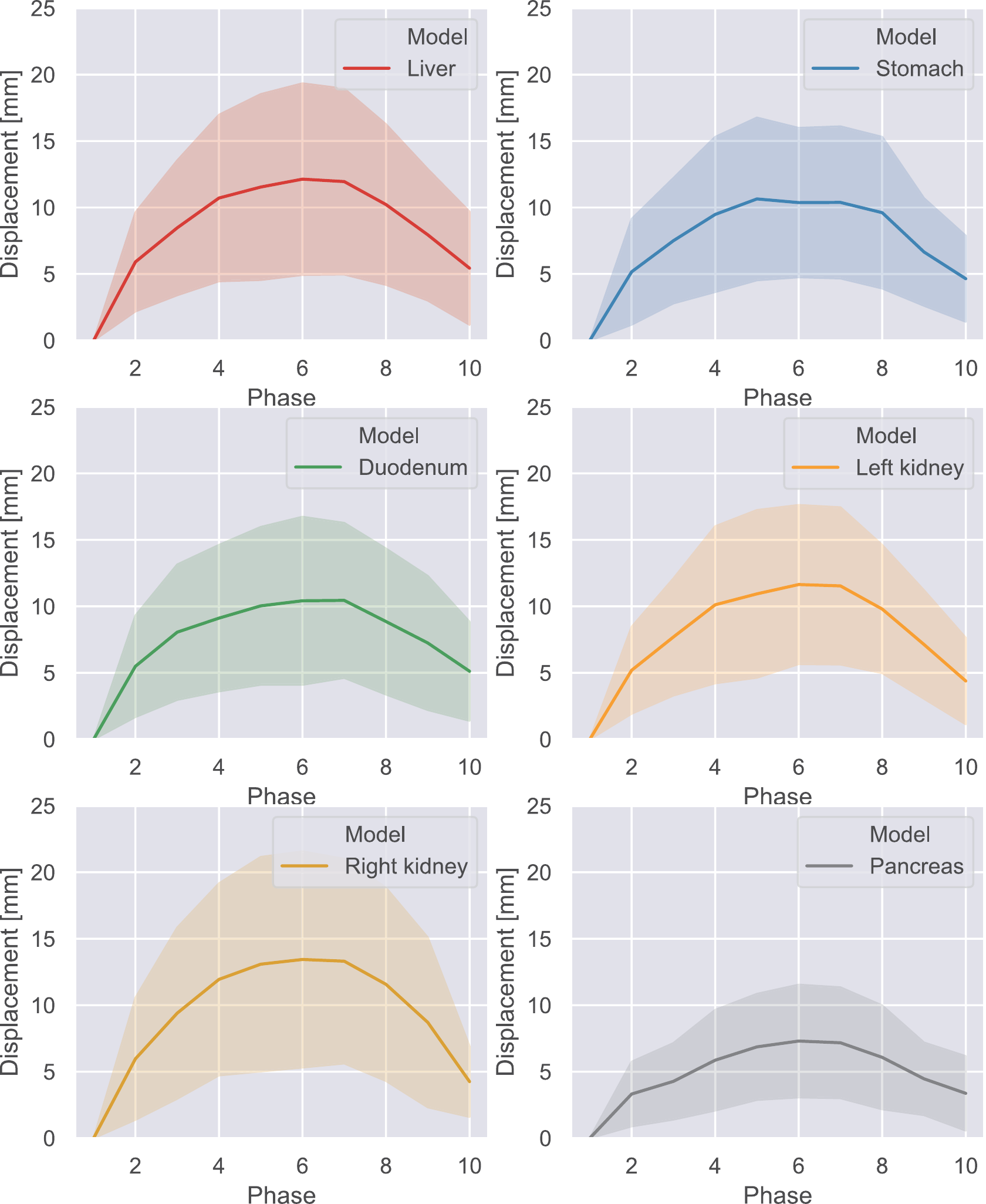}
    \caption{Statistical motion dynamics of five abdominal organs and the GTV of pancreatic cancer. The means and standard deviations of the corresponding vertices are plotted in the graphs.}
    \label{fig:8}
\end{figure}    

\begin{figure*}[t]
  \centering
  \includegraphics[width=170mm]{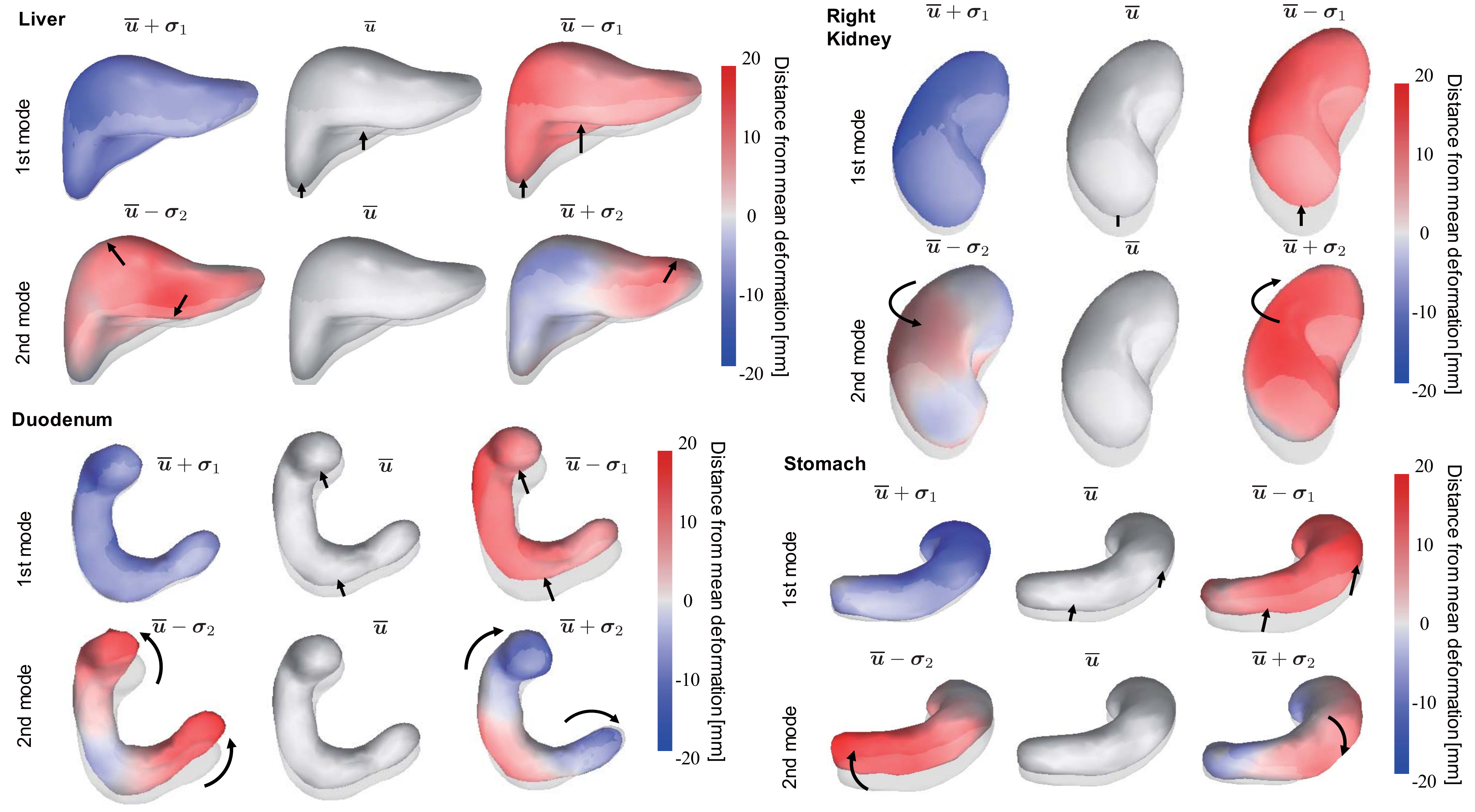}
    \caption{Statistical deformation representation of the liver, stomach, duodenum, and right kidney. Deformation variations corresponding to the first two eigenvalues from the registered models are visualized. The color map shows the signed distance from the mean deformation.}
    \label{fig:9}
\end{figure*}

Table \ref{table:2} shows the quantitative comparison results of the DMR algorithms for each of the five organs. LDSM and LSM achieved a significantly smaller mean distance and Hausdorff distance with an error of less than 1 mm, outperforming PWA and LDDMM in terms of matching volumetric regions. Regarding the Laplacian of the displacement field, LDSM obtained smaller values than LSM, indicating that smooth deformation that reduces unstable surface matching can be performed by LDSM. In DMR, the accuracy of shape representation and smooth deformation from the template is a trade-off. Using the proposed LDSM method,  the Hausdorff distance is less than 1 mm, and the Laplacian of displacement is better than that pf LSM. Based on these results, we selected the LDSM for constructing SMDM of the five organs.

\begin{table}
  \begin{center}
      \caption{Quantitative comparison results: mean distance (MD), Hausdorff distance (HD), mean and maximum Laplacian of the displacement (LD), and Dice similarity coefficient (DSC) of the DMR algorithms for the five organs.}
      \label{table:2}
  \scalebox{0.7}{
  \begin{tabular}{cccccc}
  \hline
  \multirow{2}{*}{Liver} & \multicolumn{4}{c}{Methods}    \\ \cline{2-5} 
                         & PWA                & \multicolumn{1}{c}{LDDMM} & LSM               & LDSM              \\ \hline
  MD {[}mm{]}            & 1.6 (1.1 -- 2.4)      & 0.5 (0.3 -- 0.7)             & 0.2 (0.1 -- 0.3)     & 0.2 (0.1 -- 0.3)     \\
  HD {[}mm{]}            & 10.0 (4.7. -- 28.7)   & 3.6 (1.3 -- 14.8)            & 0.9 (0.4 -- 1.9)     & 1.1 (0.5 -- 2.2)     \\
  LD (mean) {[}mm{]}     & 2.3 (1.0 -- 4.6)      & 1.1 (0.8 -- 1.6)             & 1.5 (1.0 -- 2.2)     & 1.2 (0.8 -- 1.7)     \\
  LD (max) {[}mm{]}      & 12.0 (4.6 -- 26.9)    & 6.8 (3.6 -- 13.3)            & 8.8 (4.4 -- 20.6)    & 7.0 (3.7 -- 15.2)    \\ 
  DSC [\%]               & 95.7 (92.4 -- 97.2)   & 98.0 (96.1 -- 98.6)          & 98.1 (96.1 -- 98.7)  & 98.1 (96.1 -- 98.7)  \\ \hline
  \end{tabular} 
  }

  \scalebox{0.7}{
  \begin{tabular}{cccccc}
      \hline
      \multirow{2}{*}{Stomach} & \multicolumn{4}{c}{Methods}    \\ \cline{2-5} 
                             & PWA                    & \multicolumn{1}{c}{LDDMM}     & LSM                   & LDSM              \\ \hline
      MD {[}mm{]}            & 1.4 (0.9 -- 2.1)       & 0.5 (0.2 -- 1.1)              & 0.2 (0.1 -- 0.3)      & 0.2 (0.1 -- 0.4)     \\
      HD {[}mm{]}            & 7.3 (3.7 -- 14.9)      & 3.1 (1.3 -- 10.4)             & 0.8 (0.3 -- 1.9)      & 0.9 (0.4 -- 1.8)     \\
      LD (mean) {[}mm{]}     & 2.1 (0.8 -- 5.0)       & 1.1 (0.7 -- 2.0)              & 1.3 (0.8 -- 2.5)      & 1.1 (0.6 -- 2.0)     \\
      LD (max) {[}mm{]}      & 8.7 (3.6 -- 43.2)      & 6.3 (3.0 -- 10.8)             & 7.7 (3.9 -- 15.7)     & 6.6 (2.9 -- 12.2)    \\ 
      DSC [\%]               & 92.6 (89.4 -- 95.6)    & 96.5 (95.3 -- 97.5)           & 97.2 (96.0 -- 98.1)   & 97.0 (95.7 -- 98.2)  \\ \hline
  \end{tabular}
  }

  \scalebox{0.7}{
  \begin{tabular}{cccccc}
      \hline
      \multirow{2}{*}{Duodenum} & \multicolumn{4}{c}{Methods}    \\ \cline{2-5} 
                             & PWA                    & \multicolumn{1}{c}{LDDMM} & LSM                   & LDSM              \\ \hline
      MD {[}mm{]}            & 1.2 (0.7 -- 2.9)       & 0.5 (0.3 -- 1.4)          & 0.2 (0.1 -- 0.5)      & 0.2 (0.1 -- 0.5) \\
      HD {[}mm{]}            & 6.7 (2.9 -- 23.0)      & 3.6 (0.8 -- 13.1)         & 0.8 (0.3 -- 4.2)      & 0.9 (0.4 -- 5.0) \\
      LD (mean) {[}mm{]}     & 1.5 (0.6 -- 2.6)       & 1.0 (0.6 -- 2.1)          & 1.2 (0.6 -- 2.9)      & 1.0 (0.6 -- 2.3) \\
      LD (max) {[}mm{]}      & 6.6 (2.2 -- 16.3)      & 6.9 (3.0 -- 26.8)         & 7.8 (2.9 -- 28.6)     & 7.2 (2.9 -- 22.5) \\ 
      DSC [\%]               & 86.9 (54.1 -- 92.6)    & 93.1 (70.6 -- 96.4)       & 94.2 (69.0 -- 97.4)   & 94.2 (72.2 -- 97.2) \\ \hline
  \end{tabular}
  }

  \scalebox{0.7}{
  \begin{tabular}{cccccc}
      \hline
      \multirow{2}{*}{Left kidney} & \multicolumn{4}{c}{Methods}    \\ \cline{2-5} 
                             & PWA                    & \multicolumn{1}{c}{LDDMM} & LSM                   & LDSM              \\ \hline
      MD {[}mm{]}            & 0.8 (0.6 -- 1.2)       & 0.3 (0.2 -- 0.9)          & 0.1 (0.1 -- 0.2)      & 0.1 (0.1 -- 0.3) \\
      HD {[}mm{]}            & 4.8 (2.7 -- 9.4)       & 2.8 (0.7 -- 16.4)         & 0.5 (0.2 -- 1.8)      & 0.7 (0.2 -- 2.1) \\
      LD (mean) {[}mm{]}     & 1.7 (0.7 -- 4.1)       & 0.9 (0.5 -- 2.2)          & 1.2 (0.6 -- 2.9)      & 1.0 (0.5 -- 2.3) \\
      LD (max) {[}mm{]}      & 7.5 (2.6 -- 21.7)      & 6.9 (3.1 -- 15.4)         & 7.8 (3.0 -- 24.6)     & 7.2 (2.6 -- 21.7) \\ 
      DSC [\%]               & 95.9 (90.0 -- 96.9)    & 97.2 (95.9 -- 98.4)       & 97.4 (96.0 -- 98.4)   & 97.4 (95.9 -- 98.5) \\ \hline
  \end{tabular}
  }

  \scalebox{0.7}{
  \begin{tabular}{cccccc}
      \hline
      \multirow{2}{*}{Right kidney} & \multicolumn{4}{c}{Methods}    \\ \cline{2-5} 
                             & PWA                    & \multicolumn{1}{c}{LDDMM} & LSM                   & LDSM              \\ \hline
      MD {[}mm{]}            & 0.8 (0.6 -- 1.0)       & 0.3 (0.2 -- 1.0)          & 0.1 (0.1 -- 0.2)      & 0.1 (0.1 -- 0.3) \\
      HD {[}mm{]}            & 4.3 (2.4 -- 9.1)       & 2.9 (0.5 -- 13.1)         & 0.4 (0.2 -- 0.9)      & 0.6 (0.3 -- 2.0) \\
      LD (mean) {[}mm{]}     & 1.6 (0.7 -- 3.1)       & 0.8 (0.4 -- 1.6)          & 1.0 (0.5 -- 2.0)      & 0.8 (0.5 -- 1.6) \\
      LD (max) {[}mm{]}      & 8.6 (2.7 -- 20.9)      & 6.9 (2.7 -- 14.5)         & 8.1 (2.9 -- 23.1)     & 7.0 (2.7 -- 21.6) \\ 
      DSC [\%]               & 96.2 (94.8 -- 97.3)    & 97.4 (96.3 -- 97.9)       & 97.7 (96.6 -- 98.6)   & 97.6 (96.4 -- 98.4) \\ \hline
  \end{tabular}
  }

  \end{center}

\end{table}

\subsection{Multi-organ deformation analysis}
So far, no study has investigated the impact of inter- and intra-subject variation on abdominal multi-organ deformation. Our DMR framework can directly provide a statistical representation of the registered organ models $M_{t}^{(k)}$, which can then generate the mean and variation of deaeration deformation between subjects. 

\subsubsection{Statistical motion dynamics}

Figure \ref{fig:8} shows the statistical motion dynamics with deformation computed from registered organ models for ten time phases in one respiratory cycle. The mean displacement for all corresponding vertices are visualized as the centerline, and the standard deviation is depicted as a colored band. The graph shows that the mean and standard deviation of the displacement at the end-expiration phase is $12.1 \pm 7.2$ mm for the liver, $10.3 \pm 5.5$ mm for the stomach, $10.2 \pm 5.9$ mm for the duodenum, $11.4 \pm 6.2$ mm for the left kidney, $13.8 \pm 8.4$ mm for the right kidney, and $7.6 \pm 4.2$ mm for the GTV of the pancreatic cancer. The standard deviation is relatively large compared with the magnitude of the displacement. This indicates that there are large individual variations in respiratory motion and organ deformation. Hence, to estimate the 3D tumor region with only a mean deformation model would be difficult. 

\begin{figure*}[t]
  \centering
  \includegraphics[width=180mm]{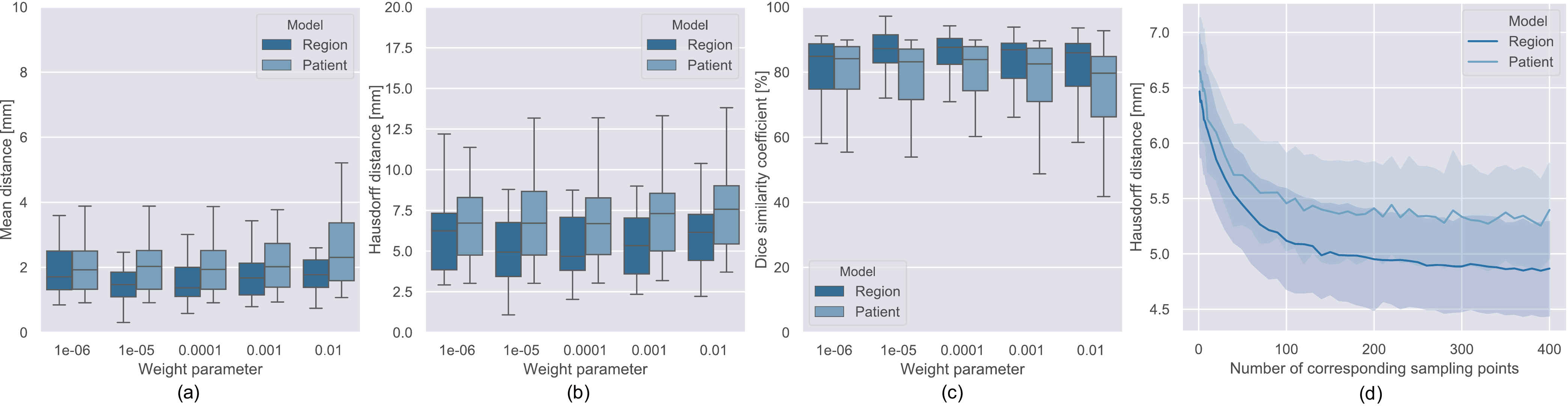}
  \caption{Comparison of prediction performance between per-patient- and per-region-based learning. (a) mean distance, (b) Hausdorff distance, (c) Dice similarity coefficient and (d) estimation error of tumor localization with respect to the number of sampling points. }
  \label{fig:10}
\end{figure*}


\subsubsection{Statistical deformation model}

Figure \ref{fig:9} shows the deformation modes that correspond to the first two eigenvalues of the registered organ models. The eigenvalues and eigenvectors were computed from the set of displacement vectors of all vertices based on singular value decomposition. The central figures show the mean shape and mean displacement. The translucent shape is the end-inspiration phase ($t=1$), and the opaque shape is the end-expiration phase ($t=6$). The left/right images were generated by changing the weights to plus/minus $\sigma$, which is twice the square root of the eigenvalues. The supplemental movies visualize the sequential motion expression by interpolating the end-inspiration and end-expiration phases. The color map shows the signed distance from the mean deformation. 

The types of variety of motion and deformation can be characterized according to their morphological properties as follows:

\begin{itemize}
    \item The first eigenvector mainly encompasses variation in the scale of deformation, which indicates that individual difference is large during the respiratory cycle.
    \item The second eigenvector is associated with the directions and rotations of deformation. Interestingly, the rotation axis and direction of rotation differ for each organ.
\end{itemize}

We also confirmed that the subspace representation using two eigenvectors explains 96.1\% of the total deformation variation.

\subsection{Deformation reconstruction performance}



The aim of the next experiment was to investigate the prediction performance and characteristics of per-patient- and per-region-based learning models. For the experimental setup, 100 corresponding vertices were randomly sampled from each organ model, and a total of 500 vertices were used as multi-organ shape features, meaning that the dimensionality of the feature vector $\Vec{x}$ was 3,000. The displacement vectors of the GTV $\Vec {y}$ were calculated using Eq. (1) through leave-one-out cross-validation. The per-patient- and per-vertex-based learning models were evaluated using the mean distance, Hausdorff distance, and Dice similarity coefficient between the estimated and ground truth regions of the pancreatic cancer. Figure \ref{fig:10} shows box plots of the three error metrics for different weight parameter values. Significant differences were found for all the error metrics when using $10^{-5}, 10^{-4}, 0.001, 0.01$ for the weight parameters by one-way analysis of variance (ANOVA; p < 0.05 significance level). The minimal estimation error was $6.7 \pm 2.6$ mm for per-patient-based learning, and $4.8 \pm 2.3$ mm for per-vertex-based learning, which shows that per-vertex-based learning improved the estimation performance by 28.4\%. 


The characteristics of the multidimensional feature sets depend on the complexity of the tumor localization problem and the number of data sets. These are both important factors affecting the estimation performance and the calculation cost. Therefore, we investigated the relationship between the estimation error of tumor localization and the number of sampling points $N$. A mean estimation error was calculated for ten trials of random sampling from five organs while increasing $N$ from 1 to 400. The weight parameter of the reproducing kernel was set to $3.0 \times 10^{-5}$, which results in good estimation performance. 

Figure \ref{fig:10}(d) shows the transition of the prediction performance of the two models. In the graph, per-region-based learning model shows consistently better prediction performance regardless of the number of sampling points. The estimation error decreased as the number of sampling points increased, and the error tends to converge over around 300 points. In the proposed per-vertex-based learning model, an average of 4.8 mm estimation error was achieved in the case of $N = 300$, which is a performance that is similar to the previous setting, which used 500 sampling points.

\begin{figure*}[t]
  \centering
  \includegraphics[width=180mm]{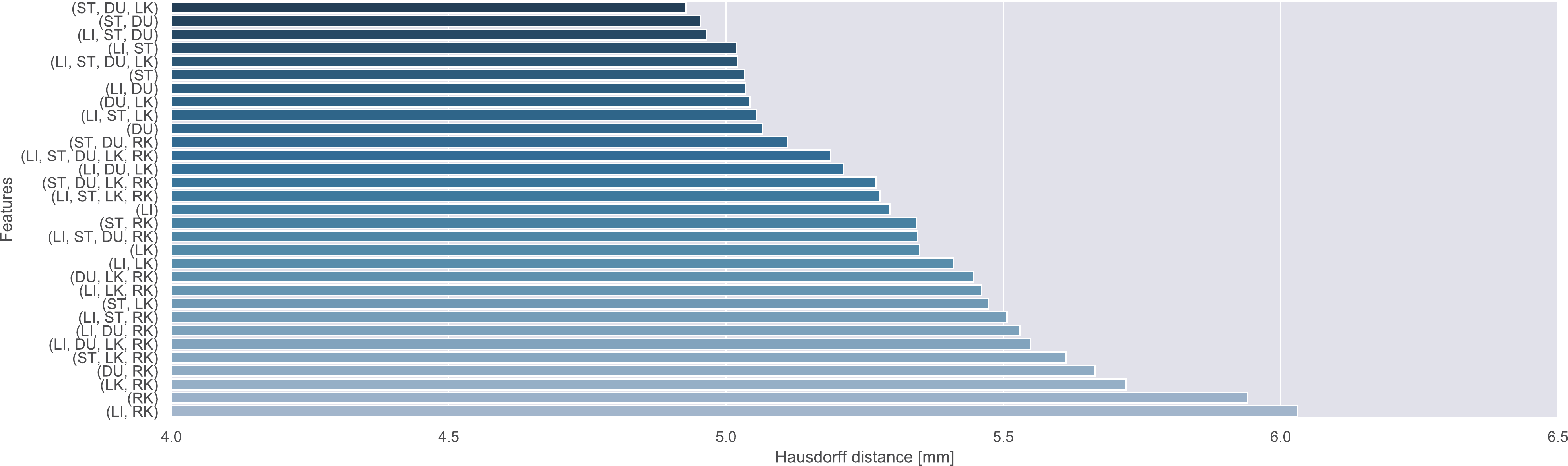}
  \caption{Multi-organ shape features and prediction performance. LI: liver, ST: stomach, DU: duodenum, LK: left kidney, and RK: right kidney.}
  \label{fig:12}
\end{figure*}

\begin{figure}[t]
  \centering
  \includegraphics[width=88mm]{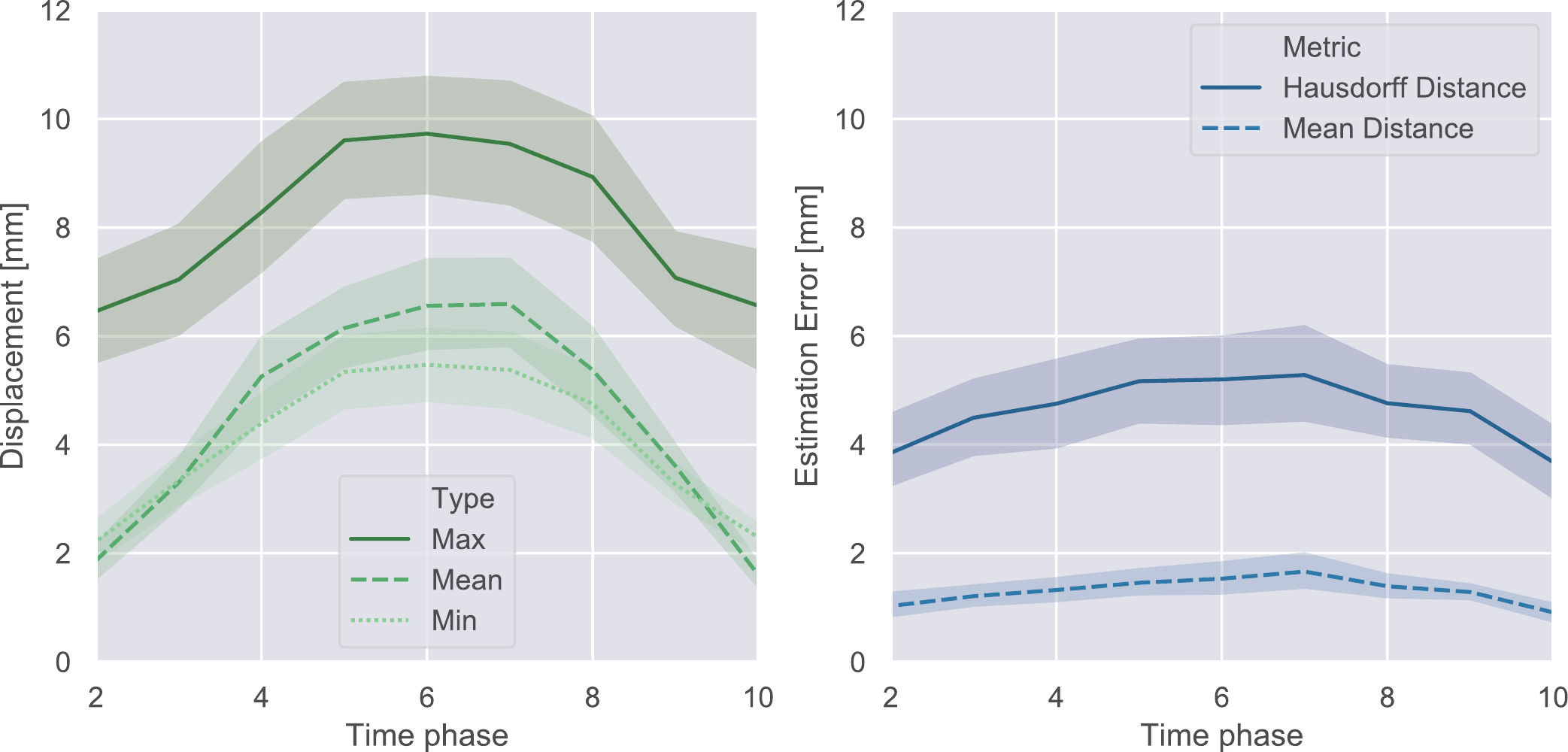}
  \caption{Inter-patient variation of the GTV's motion with local deformation and estimation errors. (a) Mean, maximum, and minimum paths over one respiratory cycle, (b) The estimation error for ten time phases. }
  \label{fig:13}
\end{figure}

\subsection{Multi-organ shape features}


The investigation of effective feature sets in respiratory motion analysis is an important topic not only for reproducing kernels but for deep learning applications. The next experiments confirm the relationship between 31 feature sets (all combinations of the five organs) and estimation accuracy in the prediction of pancreatic cancer deformation. This experiment also compares the results of the performance of the proposed multidimensional features and that of other regression approaches with a single organ or low-dimensional features. The number of sampling points was fixed to 300 based on the results of the previous experiments. 

Figure \ref{fig:12} shows the median of the Hausdorff distance sorted in ascending order for 31 feature sets. These results suggest the following: 

\begin{itemize}
    \item Features from smaller organ sets show better estimation performance than ones sampled from all five organs.
    \item Estimations using only one organ tend to increase estimation error. Specifically, the right kidney leads to poor estimation performance.
    \item The stomach, duodenum, and left kidney are the best motion descriptors for estimating the deformable region of pancreatic cancer. 
\end{itemize}

These findings suggest the validity of using the features of multiple neighboring organs rather than features sampled from the entire abdominal area. Shape features from the liver perform worse despite the fact that this organ is relatively close to the pancreas, but they are good candidates for feature descriptors when the contours of the stomach and duodenum are not available. 

\subsection{Estimation performance on motion dynamics}

The goal of the motion/deformation analysis in this paper was to investigate the estimation performance of pancreatic cancer using multi-organ shape features. We analyzed the estimation error of GTV for ten time phases of the 4D-CT dataset, as shown in Fig. \ref{fig:13}. The analysis of the multi-organ feature sets show that, the stomach, duodenum, and left kidney were used for this prediction. The number of sampling points and parameters for the reproducing kernels were the same as in the previous experiments. Figure \ref{fig:13}(a) shows the inter-patient variation of the GTV's motion with local deformation. The mean, maximum, and minimum path in one respiratory cycle are plotted as three lines, and each colored bandwidth is the standard deviation for 25-patient data. Because the displacement is calculated per vertex for the pancreas model, the difference between the maximum and minimum paths represents a local deformation exceeding 5 mm on average. The overall path also includes considerable variations with over 10 mm differences in individual respiratory motion. This can also be confirmed in the full bandwidth of Fig. \ref{fig:8}. 

The transition of the estimation error for ten time phases using the designed reproducing kernel is plotted in Fig. \ref{fig:13}(b). Despite the inter-patient variation of motion and deformation observed in Fig. \ref{fig:13}(a), the mean $\pm$ standard deviation of the distance was 1.2 $\pm$ 0.7 mm and that of the Hausdorff distance was 4.2 $\pm$ 2.3 mm. Both errors remained small throughout the respiration. These results suggest that the GTV can be adequately localized by the shape features of the surrounding organs, specifically the stomach, duodenum, and left kidney, even if the pancreas is not directly detected.

\section{Discussion}
    
To our knowledge, this study is the first to build an SMDM, a statistical multi-organ deformation library of five abdominal organs that includes inter- and intra-patient shape variations. Image-based deformable registration \cite{Sotiras13}\cite{Oh17} is a popular approach for deformed bodies, however, matching multi-organ regions tends to result in large registration errors, especially around the organ boundaries. The potential of DMR has recently been rediscovered \cite{Rigaud19}\cite{Magallon18}\cite{Nakamura19}, and the proposed LDSM for each organ address problems with matching organs with rotational components and sliding boundaries. Moreover, it achieved stable registration with a Hausdorff distance error of less than 1 mm. We note that this result outperforms the registration error of 3.1--3.3 mm reported for the SSM of CTV of the cervix--uterus and bladder in recent work \cite{Rigaud19}. 


To clarify the focus of this research, the spatial displacement of the internal structures of the mesh was considered to be outside the scope of this paper. This is because the assumed application of SMDM is radiotherapy planning, in which the 3D contour (i.e., surface) of the GTV and OARs must be estimated. Because the developed library uses a tetrahedral mesh in the DMR and can express internal deformation, we consider that further modeling and evaluation of internal structures would be possible in future work.

In this experiment, the kernel-based modeling approach was used as a nonlinear deformation regression for a limited number of 4D-CT data. Despite this study's limited data size, the results showed that stable estimation could be achieved throughout the respiration cycle. These findings and the developed database will be useful for determining the planning target volume of pancreatic cancer from the available features of the surrounding organs. Some recent studies report pixel-to-shape techniques, i.e., 3D shape/deformation reconstruction from a single 2D image \cite{Wang19}\cite{Wu19}\cite{Nakao17}. Combining such 3D shape reconstruction techniques, SMDM, and extracted multi-organ feature sets,  would be an interesting approach for real-time tumor localization from X-ray images.Our future work includes application of the developed multi-organ library to markerless radiation for tumor-tracking radiotherapy.


%

\section{Conclusion}
In this paper, we introduced a multi-organ deformation library and its application to pancreatic cancer localization based on the shape features of multiple organs. The statistical multi-organ motion/deformation library of the stomach, liver, left and right kidneys, and duodenum was generated by the DMRs of their organ meshes generated from 4D-CT images (250 volumes). The proposed LDSM method achieved stable registration with a Hausdorff distance error of less than 1 mm. Per-region-based deformation learning using a reproducing kernel was also proposed to predict the displacement of pancreatic cancer for ART. The experiment results show that the proposed concept better estimates, and achieves a clinically acceptable estimation error for mean distance (1.2 $\pm$ 0.7 mm) and the Hausdorff distance (4.2 $\pm$ 2.3 mm) throughout the respiratory motion. 


\bibliographystyle{IEEEtran}

\ifCLASSOPTIONcaptionsoff
  \newpage
\fi






\vfill

\end{document}